# Parasitic Egg Detection and Classification in Low-cost Microscopic Images using Transfer Learning

Thanaphon Suwannaphong[a,*], Sawaphob Chavana[b], Sahapol Tongsom[b], Duangdao Palasuwan[c], Thanarat H. Chalidabhongse[b] and Nantheera Anantrasirichai[d]

[a]*Department of Engineering Mathematics, University of Bristol, Bristol, UK*
[b]*Department of Computer Engineering, Chulalongkorn University, Thailand*
[c]*Research Group on Applied Computer Engineering Technology for Medicine and Healthcare, Chulalongkorn University, Thailand*
[d]*Visual Information Laboratory, University of Bristol, Bristol, UK*

ABSTRACT

Intestinal parasitic infection leads to several morbidities to humans worldwide, especially in tropical countries. The traditional diagnosis usually relies on manual analysis from microscopic images which is prone to human error due to morphological similarity of different parasitic eggs and abundance of impurities in a sample. Many studies have developed automatic systems for parasite egg detection to reduce human workload. However, they work with high quality microscopes, which unfortunately remain unaffordable in some rural areas. Our work thus exploits a benefit of a low-cost USB microscope. This instrument however provides poor quality of images due to limitation of magnification (10×), causing difficulty in parasite detection and species classification. In this paper, we propose a CNN-based technique using transfer learning strategy to enhance the efficiency of automatic parasite classification in poor-quality microscopic images. The patch-based technique with sliding window is employed to search for location of the eggs. Two networks, AlexNet and ResNet50, are examined with a trade-off between architecture size and classification performance. The results show that our proposed framework outperforms the state-of-the-art object recognition methods. Our system combined with final decision from an expert may improve the real faecal examination with low-cost microscopes.

## 1. Introduction

Intestinal parasitic infection leads to several morbidities such as diarrhea digestive disorders, malnutrition, anemia or even death to humans worldwide, especially in tropical countries [1]. There are more than 100 different species of human intestinal parasites [2] that can grow with a considerably rapid rate of 200,000 eggs re-produced daily [3]. Indubitably, over 415,000 human deaths due to parasite infestation are reported annually [4]. The conventional methods to diagnose intestinal parasitic infection depend exclusively on medical technicians in microscopic examinations of faecal samples. The potential problems are morphological similarity of parasites and abundance of impurities in a sample, causing a difficulty to manually inspect different types of parasite eggs via a microscope [5, 3]. This thus requires extensive training to gain adequate expertise in diagnosis. This examination is laborious and time-consuming, averaging 8–10 mins required for an expert technician to examine one sample [4]. Also, the lack of skilled workers in this field leads to exhaustion, which means this method is prone to human errors [6]. Moreover, these conventional methods do not provide a data-sharing system and record historical data of diagnosis. For these reasons, the invention of an automated diagnostic system would become a significant contribution to assist the traditional diagnosis.

Since the development of image processing techniques and computer vision, the automated diagnosis systems have become more feasible. Many studies have implemented the systems to analyse microscopic image of faecal samples based on machine learning, e.g. support vector machine (SVM) [7, 8] and artificial neural networks (ANN) [5, 7]. The core components of these systems commonly include pre-processing, feature extraction and classification to identify species of parasitic eggs. The approaches generally consider the difference of the external features such as size, shape and smoothness of eggshell. These traditional machine learning approaches do not require vary complex designs; however, they rely heavily on the selectively extracted features. This means a tremendous effort is needed for tuning the features in feature extraction stage.

In the last decade, as computer performance and quantity of available image datasets have both increased, the deep learning-based algorithms have rapidly become more popular [9]. Deep learning shows high efficiency to tackle problems in many different domains, such as text recognition, computer-aided diagnosis, face recognition and drug discovery [10]. In the parasite egg detection task, deep learning, especially the convolutional neural networks (CNNs), motivates novel parasite classification research by its promising performance and speed in object detection [4, 6, 11, 12, 13, 14]. CNNs have an advantage to avoid manual feature extraction by using its ability to learn the relevant features automatically from a large amount of data that represents the desired behaviour of the data [15]. CNNs have demonstrated high accuracy in numerous different tasks of pathogen detection such as malaria, tuberculosis, and intestinal parasite [6].

*Corresponding author
✉ ic19511@bristol.ac.uk (T. Suwannaphong);
n.anantrasirichai@bristol.ac.uk (N. Anantrasirichai)
ORCID(s):

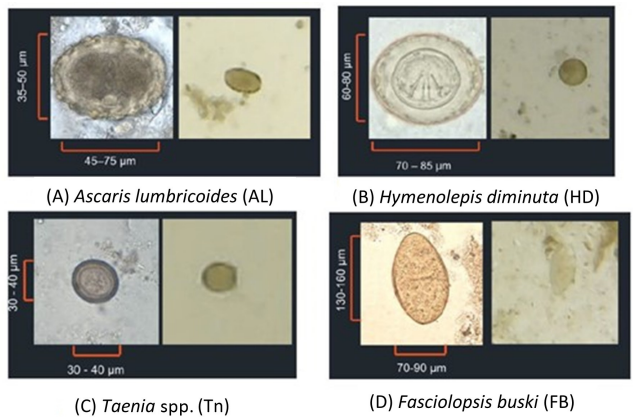

**Figure 1:** The comparison of parasitic egg images from (left) good quality microscope and (right) USB microscope

With this approach, the intestinal parasite diagnosis can be independent of the experts, save time, increase sensitivity, and hopefully be used on-site.

Previous automatic systems proposed for the parasite egg detection have employed high quality microscopes (1000×). These instruments are expensive and have limited availability in rural areas. Therefore, we are motivated to employ a low-cost USB microscope in this paper. It comes with low magnification (10×), generating poorer image quality as low contrast and relevantly less detail of parasite eggs as shown in figure 1. At high resolutions (high magnification), the images of different parasite species show unique characteristics and textural patterns inside the eggs (left images). This aids the CNNs to learn these expressed image patterns and assists identifying the parasite species. In contrast, the images from low magnification contain obviously fewer characteristics and pattern details of the parasitic eggs, leading to difficulty in detection and species classification.

This paper addresses these challenges by applying a transfer learning technique with pretrained CNN networks. Our proposed framework starts with image processing to enhance the image quality. We subsequently employ the transfer learning strategy and test two pretrained networks: AlexNet [16] and ResNet50 [17], to examine a trade-off between the lighter weight architecture and the better classification performance. Our detection technique is a patch-based sliding window, where the unlabelled microscopic images are separated into overlapping patches, fed to the trained model, and the location of the detected egg is where the maximum probability of being parasite egg is. Our proposed automated system based on supervised deep learning shows potential to enhance the efficiency of automatic parasite egg detection and classification in poor-quality microscopic images that have never been investigate in the state of the art. Obviously, there is a difficulty to identify parasite species from these USB microscopic images; however, our system could be a preliminary diagnostic tool before the next process for species identification can be perform for more specific species classification.

## 2. Materials and methods

### 2.1. Parasite eggs in USB microscopic images

Our dataset contains 10× magnification microscopic images with a resolution of 640×480 pixels. Four different types of parasitic eggs present: 67 images of *Ascaris lumbricoides* (AL), 27 images of *Hymenolepis diminuta* (HD), 32 images of *Fasciolopsis buski* (FB) and 36 images of *Taenia* spp. (Tn). The images were obtained from Chulalongkorn University, Thailand, and were labelled by experts. The limitation of the low magnification causes the difficulty in species identification because of lack of details of the parasitic eggs present in the images as shown in figure 3. The AL is the most common roundworm in humans [2]. The egg appearance is generally round but slight oval shapes are also found (Figure 3 (a)) with varied degrees of ellipticity. The HD eggs under the USB microscope look very similar to the AL eggs but the HD eggs are generally more circular than the AL eggs (Figure 3 (b)). Figure 3 (c) shows a large parasitic egg, *Fasciolopsis buski*, of which the detail showed is very low. The parasitic egg in figure 3 (d) shows common characteristics of *Taenia* eggs but the egg detail is not enough to specify the *Taenia* species. So, the experts classify these eggs as *Taenia* spp.

### 2.2. Data preparations

Greyscale conversion and contrast enhancement are performed before processing the patch overlapping and data augmentation. The greyscale conversion reduces the depth of the input image from three channels (RGB) to one channel (greyscale), diminishing the computational complexity of the training process. The contrast enhancement improves visualisation of the low magnification microscopic images. This aids the CNN model to detect the low-level features, like edges and curves, and contributes to detection performance of higher-level features, like characteristics of parasitic eggs.

#### 2.2.1. Overlapping patches

Each microscopic image is divided into patches, allowing the model to characterise the whole image by analysing local areas. The largest parasite egg is approximately 80×20 pixels. The patch size is thus defined to be 100×100 pixels so that all types of parasitic egg can be entirely encapsulated. The positions of patches overlap by four fifths of the patch size as shown in figure 4. This empirical choice provides a good prediction result when merging probability of all patches together to reconstruct the probability map corresponding the input microscopic image. To label the training dataset, the patch that contains a parasite egg is labelled as an egg patch, whilst the patch without parasite egg is labelled as the background.

#### 2.2.2. Data augmentation

Generally each microscopic image contains only 1-3 eggs, resulting highly imbalanced training dataset as there are numerous background patches. Data augmentation is therefore employed to increase the number of egg patches and to bal-ance the number of patches in each class. This approach also ensures that data points are positioned to prevent overfitting. The augmentation increases the variance of the training dataset, which makes the model invariant to location and orientation of eggs presenting in each patch. In this work, we generate more egg patches by (i) randomly flipping horizontally and vertically, (ii) randomly rotating between angle of 0 to 160 degree, and (iii) randomly shifting every 50 pixels horizontally and vertically around the egg. This data augmentation technique increases the egg patches to approximately 10,000 patches per each egg type. We randomly select 10,000 background patches so that the numbers are balanced.

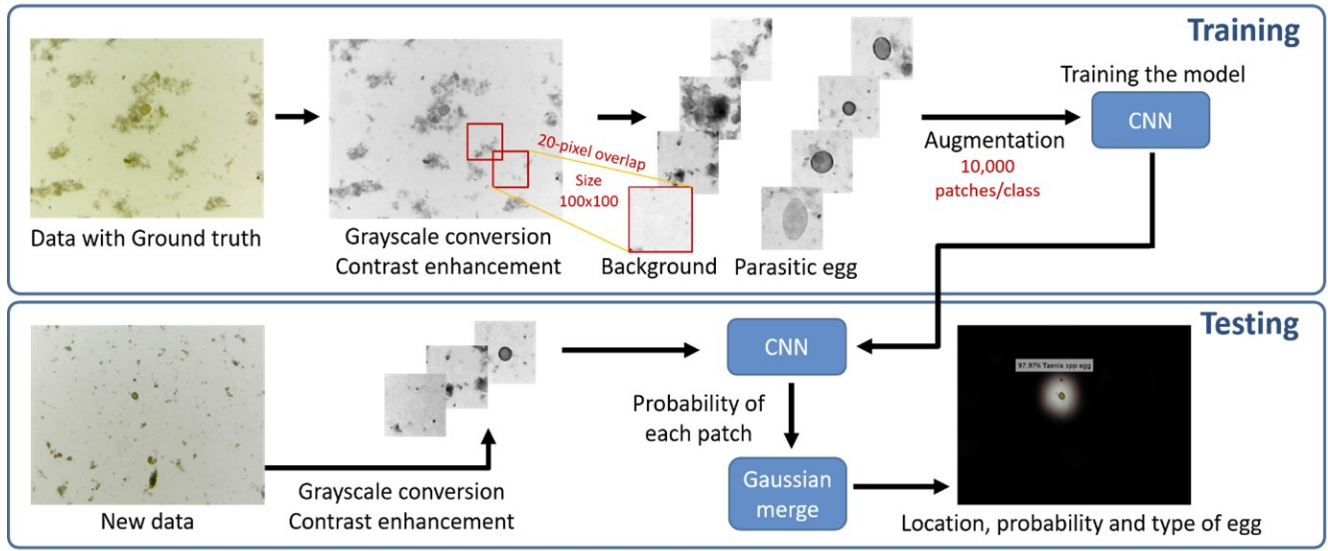

**Figure 2:** Diagram of the proposed framework for training and testing CNN model

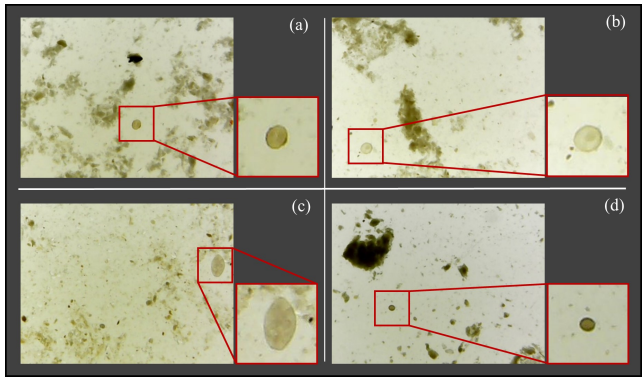

**Figure 3:** USB microscopic images of 4 types of parasitic eggs (a) *Ascaris lumbricoides* (b) *Hymenolepis diminuta* (c) *Fasciolopsis buski* and (d) *Taenia* spp.

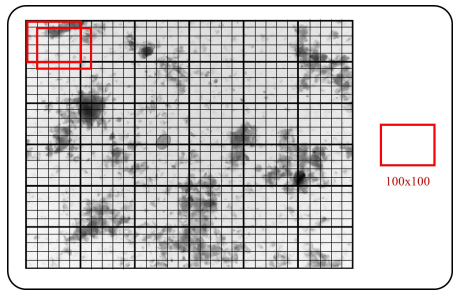

**Figure 4:** Patch overlapping technique

### 2.3. Methods with transfer learning

We employ a transfer learning strategy by fine-tuning pretrained networks. This approach is faster and easier than training a network with randomly initialised weights from scratch. Parameters and features of these networks have been learnt from a very large dataset of natural images thereby being applicable to numerous specific applications. The last two layers are replaced with a fully connected layer and a softmax layer to give a classification output related to five classes: 4 parasite egg types and background debris. The learning rates of the new layers are defined to be faster than the transferred layers.

Two pretrained networks are tested, i.e. AlexNet [16] and ResNet50 [17], in order to examine a trade-off between network size and classification performance. AlexNet is a pioneer model that has significantly contributed to CNN performance for object recognition, whilst ResNet50 is more modern, deeper architecture and generally gives better performance on image classification.

The prepared patches as described in the previous step are resized to the input size required for each network, i.e. 227×227 and 224×224 pixels for AlexNet and ResNet50, respectively. We randomly choose 30 percent of the training patches for validation in the training process. The initial learning rate, mini-batch size, and maximum epochs are varied to find the optimal setting. The data was shuffled every epoch to avoid creating a bad mini batch which is a poor representation of the overall dataset. We select the best model from the first model that provided the lowest validation loss

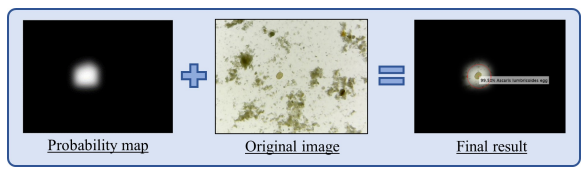

**Figure 5:** An example of a prediction result of the proposed method

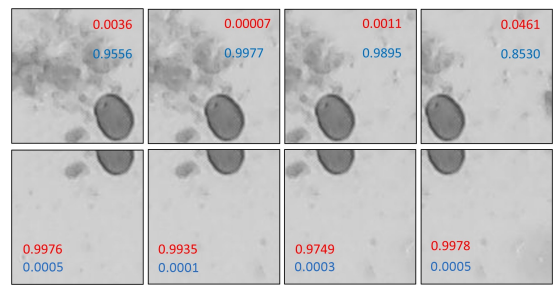

**Figure 6:** Examples of testing result for egg patches. Red values represent the probability of the patch to be background debris. Blue values represent the probability of the patch to be an AL egg

to avoid overfitting.

### 2.4. Prediction process

The testing microscopic images are processed with greyscale conversion, contrast enhancement, and patch overlapping process usingg the same parameters as the training dataset. Each patch in an image is classified using the trained models which provide the probability of each overlapping patch to identify the type of parasite eggs or background debris. The classification result of each patch gives five probability values of being AL, HD, FB , Tn and background. The final prediction result of the whole image comes from the maximum value of these five probabilities among all overlapping patches of the whole image. Then, we construct the probability map of the whole image by merging of the probabilities of every overlapping patch with Gaussian weights ($\mu = 0$, $\sigma = 1$, where $\mu$ and $\sigma$ are the mean and the standard deviation, respectively). The location of the detected parasite egg is where the highest values of the probability map is, as shown in figure 5.

## 3. Experimental results and discussion

The dataset of each parasitic egg type is divided into training and testing with a ratio of 60:40. So, the testing dataset consists of 65 images: 27 AL, 11 HD, 13 FB , and 14 Tn images. In the fine-tuning process, two optimisers were compared: stochastic gradient descent with momentum (SGDM) and Adaptive Moment Estimation (Adam). The initial learning rate was varied: 0.001, 0.0001 and 0.00001. The mini-batch size was varied between 20, 50 and 100. For both AlexNet and ResNet50, we found that the SGDM optimiser with the initial learning rate of 0.0001, the mini-batch size of 100, and 20 epochs gave the best result (achieving the training accuracy of 93.95% and 99.80% for AlexNet and ResNet50, respectively).

The validation in the training process shows that both models can classify background patches with 100% accuracy. For the AL patches, the ResNet50 only shows one misclassified patch to be a background patch whereas AlexNet produces more misclassified patches as background as well as misclassified patches to be other types of eggs. ResNet50 identifies HD patches with no mistake while AlexNet shows 7% misclassified HD patches. The FB patches are the most difficult to identify because FB eggs have very low contrast, making it less noticeable compared to other parasitic eggs. Unsurprisingly, the AlexNet performance to classify the FB patches is relatively low compared to other types of eggs. However, the ResNet50 provides an impressive result in FB patch classification with an interestingly high accuracy of 93.50%. In the same way, ResNet50 also gives a better result than AlexNet for the Tn patches.

### 3.1. Classification results

In the testing process of classification, the true positive rate is relatively low when analysing each patch separately for both models, as reported in table 1 with an analysis type of 'patch'. Several egg patches are classified as background (the highest probability is given to the background class). Most of these patches have an egg located at the edge of the patches – in other words, only a partial egg presents in the patches as displayed in figure 6. In the detection process, the probabilities of overlapping patches are merged into the probability map of the testing microscopic image. This results in an increase in the probability of there being a parasitic egg in the egg area. As a result, the true positive rates increase when analysing the whole image (table 1 with analysis type of 'whole image'). Overall, the ResNet50 model outperforms the AlexNet one in all egg cases, measured by the accuracy, true positive and true negative rates. The true negative rates are very high for the patch analysis of both AlexNet and ResNet50 – background patches are classified as background. Unfortunately, we cannot determine the true negative rates for whole image analysis because we do not have any microscopic images containing only debris (no parasite eggs) to test the model.

Figure 7 shows some false positives in the detection results of AlexNet. These areas have smaller probability values than the correctly detected areas in other images. The ResNet50 model yields better results than the AlexNet model as shown in the confusion matrices of figure 8. Interestingly, the ResNet50 model achieves 100% accuracy of FB detection (figure 8d), which is the most challenging egg in this study due to the low contrast and lack of details inside the eggs. This can be seen via the result of the patch analysis, where the FB patches yield the lowest accuracy because almost half of FB patches are classified as background debris (figure 8c). In addition, the ResNet50 generally gives higher probability than the AlexNet for either being parasite eggs or background debris, indicating more confidence of prediction. In some images, neither AlexNet nor ResNet50 can detect any parasite eggs. However, these egg areas have sig-nificantly lower probability values of being background than other parts of the image.

**Table 1**
Testing results of AlexNet and ResNet50 models

| Models | Analysis type | Accuracy(%) | True positive rate(%) | | | | True negative rate(%) |
|---|---|---|---|---|---|---|---|
| | | | AL | HD | FB | Tn | |
| AlexNet | Patch | 96.93 | 45.45 | 34.23 | 19.92 | 53.76 | 99.23 |
| | Whole image | 87.69 | 96.30 | 81.82 | 76.92 | 85.71 | - |
| ResNet50 | Patch | 98.25 | 64.09 | 73.42 | 56.39 | 73.48 | 99.57 |
| | Whole image | 90.77 | 92.59 | 81.82 | 100.00 | 85.71 | - |

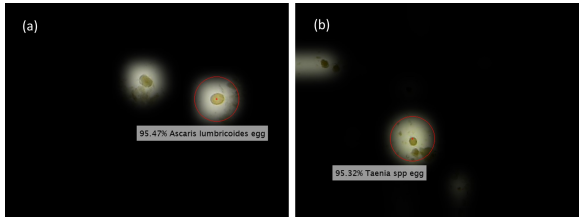

**Figure 7:** AlexNet confusion results: (a) detected AL egg correctly (red area) but also included the debris which have similar morphology as AL egg (left bright area), (b) detected Tn egg correctly (red area) but also detect debris as Tn egg with a lower probability (left dim area).

### 3.2. Comparison with state-of-the-art object detection

We compared our methods with two state-of-the-art object detection methods: Single shot detector (SSD) [18] and Faster R-CNN [19]. SSD is a very light weight network having one single shot to detect multiple objects within the image, whilst Faster R-CNN requires two shots, one for searching for regions of interest (ROI), and the other for detecting the object in each ROI using CNNs. The backbones of SSD and Faster R-CNN used here are VGG-16 and ResNet50 architectures, respectively. It is worth to mention that You Only Look Once (YOLO) [20], a state-of-the-art real-time object detection, is not suitable for our application. This is because YOLO struggles to detect small objects within the image.

Table 2 shows the precision results of our methods compared to SSD and Faster R-CNN. The precision, or positive predictive value (PPV), indicates the percentage of correctly identified eggs from all detected eggs in one particular egg type. The proposed method with the ResNet50 gives the best average precision, followed by Faster R-CNN. Our AlexNet model outperforms SSN despite smaller architecture and fewer parameters. Our method exploits a sliding window technique which works really well in this application. This is because the size of each patch can be fixed, and the aspect ratio of the objects we want to detect (parasite eggs) is not present in various shapes, unlike natural images that contain people, cars and buildings. The adaptive size of a bounding box, as offered by SSD and Fast R-CNN, does not benefit parasite egg detection.

**Table 2**
Comparison with stat-of-the-art object detection

| Models | Precision(%) | | | | Avg. |
|---|---|---|---|---|---|
| | AL | HD | FB | Tn | |
| SSD | 86.7 | 85.1 | 73.3 | 67.3 | 78.1 |
| Faster R-CNN | 92.5 | 97.3 | 98.9 | 95.8 | 96.1 |
| Our AlexNet | **96.3** | 90.0 | **100** | 85.7 | 93.0 |
| Our ResNet50 | 96.2 | 90.0 | **100** | **100** | **96.6** |

### 3.3. Discussion

The CNNs can automatically learn relevant useful features even in microscopic images with poor magnification. The transfer learning technique is a very efficient method that has the capability to produce high performance parasite egg detection within a short time. However, the transferred model requires an optimal parameter setting for the training process to yield the best result. We found that training the model with too many epochs, e.g. 50 or 100 epochs, may result in an overfitting problem, where the model provides extremely high accuracy for a training result, but the testing accuracy is extremely poor. Therefore, early stopping techniques are required or validation loss is monitored as we employ in this paper.

The proposed frameworks, especially the model based on ResNet50, achieve a satisfactory result for parasite detection task with a sufficiently high accuracy in classifying the testing images. The model has the remarkable ability to locate a parasite egg position and identify species of the parasite egg correctly even in a low-quality microscopic image. The ResNet50 model outperforms AlexNet and SSD because of its deeper architecture, leading to better handling of the variation and non-linearity of data. The deeper networks can learn complex features. Misclassified results still occur but if we combine the prediction result of the model with the final decision from an expert, this will improve the diagnosis of parasite infection disease. However, the poor-quality microscopic image with insufficient detail is still a big challenge that prevents the model from correctly discriminating between different types of parasite eggs and sample impurities. Thus, more work is still required to improve this system.

Despite a high classification performance of the model, misclassification remains. A larger pretrained network may help the model to learn more complex features of the parasite eggs. Semantic segmentation, e.g. U-Net [21, 22] might be employed to label the ground truth more precisely. This may aid the model to extract the relevant features more accurately.

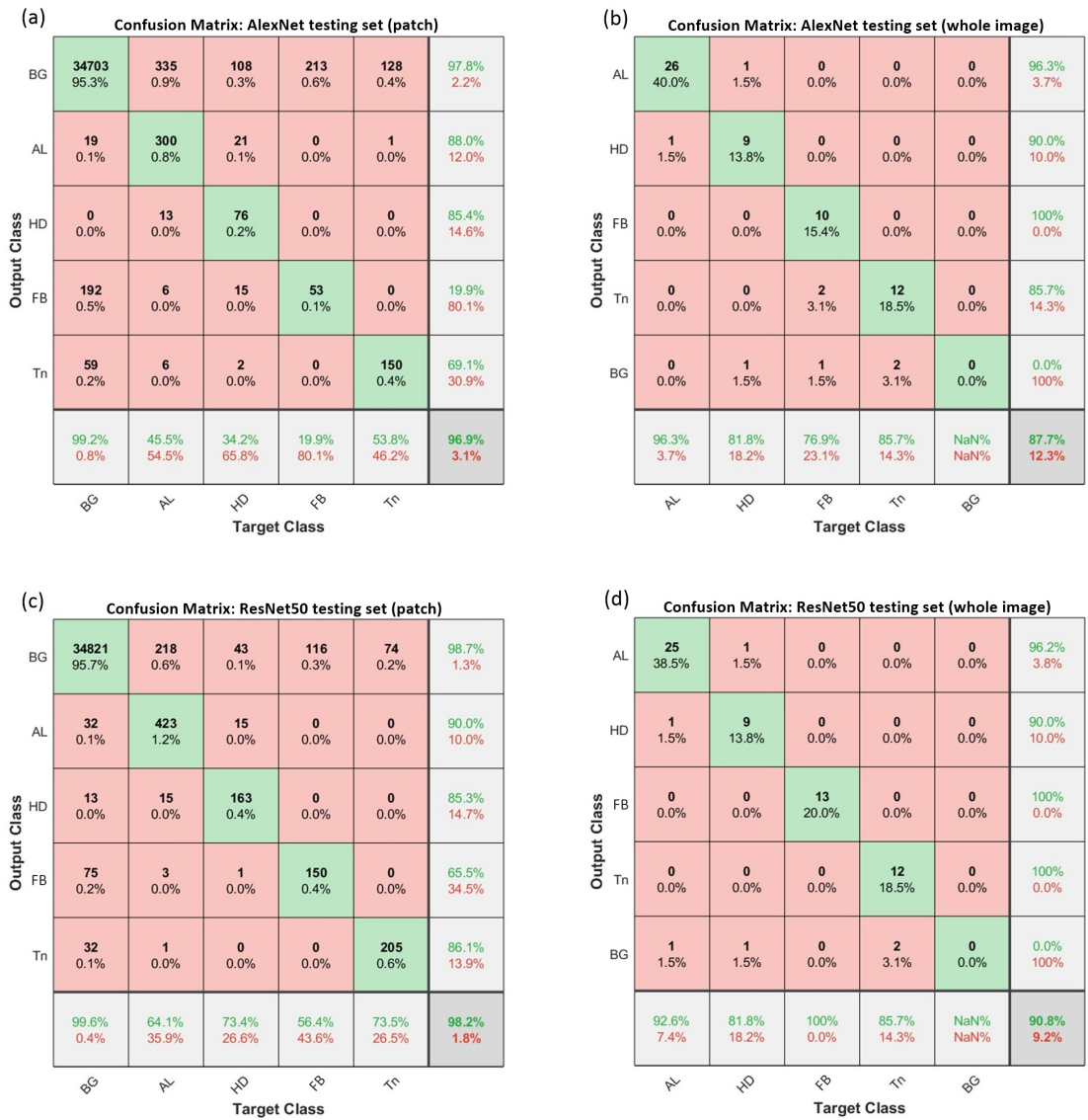

Figure 8: Confusion matrices from the testing process

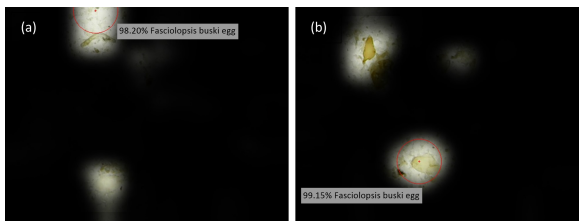

Figure 9: Examples of ResNet50 testing results for FB egg. (a) the debris at the upper bright area shows higher probability than the FB egg in the lower bright area. (b) the debris at the upper bright area shows lower probability than the FB egg in the lower bright area

## 4. Conclusions

This paper presents a deep learning technology with a transfer learning approach for an automatic system of parasite egg detection and classification during a faecal examination. The proposed CNN models show a competent ability in learning relevant features of the different parasitic eggs even in low quality images sourced from a USB microscope. Overall, our ResNet50 framework can classify the four types of parasitic eggs with high accuracy and outperforms AlexNet, SSD and Faster R-CNN. With this satisfactory result, this approach may be performed with real faecal examination with USB microscopes.

## Acknowledgment

This work was supported by Newton Fund Institutional Links [grant number 623714323] and a scholarship from Royal Thai Government under the Ministry of Higher Education, Science, Research and Innovation.